\documentclass[letterpaper]{article} % DO NOT CHANGE THIS
\usepackage[draft]{aaai2026}  % DO NOT CHANGE THIS
\usepackage{times}  % DO NOT CHANGE THIS
\usepackage{helvet}  % DO NOT CHANGE THIS
\usepackage{courier}  % DO NOT CHANGE THIS
\usepackage[hyphens]{url}  % DO NOT CHANGE THIS
\usepackage{graphicx} % DO NOT CHANGE THIS
\usepackage{natbib}  % DO NOT CHANGE THIS AND DO NOT ADD ANY OPTIONS TO IT
\usepackage{caption} % DO NOT CHANGE THIS AND DO NOT ADD ANY OPTIONS TO IT
\usepackage{algorithm}
\usepackage{algpseudocode}  % For algorithm pseudocode

\usepackage{booktabs}
\usepackage{multirow}

\usepackage{newfloat}
\usepackage{listings}
\usepackage{amsmath}
\usepackage{xcolor}         % For colored text in algorithms
\usepackage{xstring}
\usepackage{colortbl}
\definecolor{tagblue}{RGB}{0,102,204}
\definecolor{taggreen}{RGB}{0,153,0}
\definecolor{tagred}{RGB}{204,0,0}
\definecolor{backcolor}{RGB}{240,248,255} % light blue example

\usepackage[most]{tcolorbox}

\DeclareCaptionStyle{ruled}{labelfont=normalfont,labelsep=colon,strut=off} % DO NOT CHANGE THIS
\floatstyle{ruled}
\newfloat{listing}{tb}{lst}{}
\floatname{listing}{Listing}
\title{SIGMA: Search-Augmented On-Demand Knowledge
Integration for Agentic Mathematical Reasoning}
\author{
    Ali Asgarov, \;
    Umid Suleymanov, \;
    Aadyant Khatri
}
\affiliations{
    Department of Computer Science \\
    Virginia Tech \\
    \texttt{\{aliasgarov, umids, aadyant\}@vt.edu}
}

\usepackage{bibentry}
\begin{document}

\maketitle

\begin{abstract}
Solving mathematical reasoning problems requires not only accurate access to relevant knowledge but also careful, multi-step thinking. However, current retrieval-augmented models often rely on a single perspective, follow inflexible search strategies, and struggle to effectively combine information from multiple sources. We introduce \textbf{SIGMA} (\textbf{S}earch-Augmented On-Demand Knowledge \textbf{I}ntegration for A\textbf{G}entic \textbf{M}athematical re\textbf{A}soning), a unified framework that orchestrates specialized agents to independently reason, perform targeted searches, and synthesize findings through a moderator mechanism. Each agent generates hypothetical passages to optimize retrieval for its analytic perspective, ensuring knowledge integration is both context-sensitive and computation-efficient. When evaluated on challenging benchmarks such as MATH500, AIME, and PhD-level science QA GPQA, SIGMA consistently outperforms both open- and closed-source systems, achieving an absolute performance improvement of 7.4\%. Our results demonstrate that multi-agent, on-demand knowledge integration significantly enhances both reasoning accuracy and efficiency, offering a scalable approach for complex, knowledge-intensive problem-solving. We will release the code upon publication.
\end{abstract}

\begin{figure*}[ht!]
    \centering
    % \vspace{em} 
    \includegraphics[width=1\textwidth]{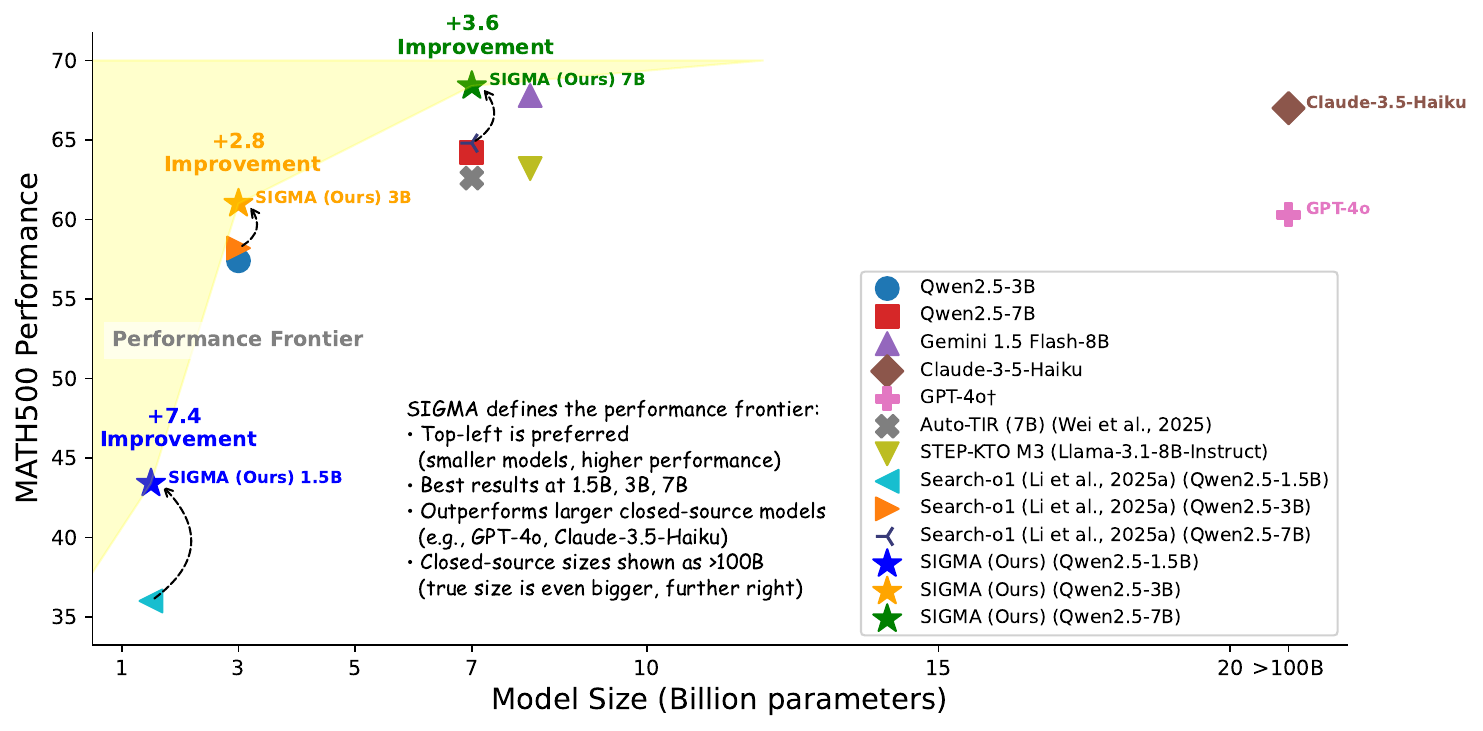}
    \vspace{-1.7em} 
    \caption{MATH500 score versus model size. SIGMA variants (stars) form a shaded performance frontier at 1.5B, 3B, and 7B parameters, with arrows indicating SIGMA’s absolute improvement over the second-best method of the same size. Closed-source models are placed at \ensuremath{>100}B on the x-axis. SIGMA outperforms several larger closed-source models, including GPT‑4o \citep{gpt_4o_system_card}, Gemini 8B \citep{GeminiTeam2025gemini25}, and Claude‑3.5‑Haiku \citep{anthropic_claude_3_5_haiku_2024}.}

    \label{fig:scatter-figure}
    \vspace{-1.5em} 
\end{figure*}

\section{Introduction}

Multi-agent systems can help with complex reasoning, but current retrieval methods usually rely on a single perspective and don’t combine different approaches effectively. Tool-Integrated Reasoning (TIR) models \citep{li2025search, Yao2023ReAct} enhance retrieval but cannot coordinate complementary strategies. \textit{Solving complex mathematical problems requires both accurate knowledge (theorems, numeric facts) and reliable step-by-step reasoning. Large models often fail because missing facts get amplified during reasoning, and single-strategy retrieval cannot consistently provide the necessary information \citep{li2025search}.}

To address these challenges, we introduce \textbf{SIGMA}, a unified framework that decomposes complex reasoning tasks across four complementary specialist agents (\textsc{Factual}, \textsc{Logical}, \textsc{Computational}, \textsc{Completeness}). Each agent independently performs targeted reasoning-search cycles, using hypothetical document enhancement to generate perspective-specific retrieval queries and search only when uncertainty arises. A lightweight moderator then synthesizes their outputs into a coherent final solution. This design ensures broad reasoning coverage without the heavy communication overhead of traditional multi-agent systems, while focusing retrieval on precise facts to break long reasoning chains and avoid the cost of monolithic, overlong chains-of-thought.

In summary, our key contributions include:

\begin{itemize}
    \item We introduce \textbf{SIGMA}, a unified framework for multi-perspective reasoning that integrates factual, logical, computational, and completeness perspectives through coordinated agents, enabling richer analysis than single-perspective methods.
    \item We propose \textbf{optimized knowledge integration} via \textit{Hypothetical Document Enhancement} and a moderator-based synthesis mechanism, tailoring retrieval to each analytical perspective and combining outputs into robust, context-sensitive reasoning.
    \item We provide \textbf{empirical validation on challenging benchmarks}, showing that SIGMA achieves consistent gains on MATH-500, AIME, and GPQA, outperforming strong open- and closed-source baselines while maintaining efficiency across model scales.
\end{itemize}

\section{Related work}

\textbf{Retrieval-augmented and multi-agent reasoning:} Large reasoning models (LRMs) like o3 \citep{OpenAI2025o3}, DeepSeek-R1 \citep{DeepSeekAI2025deepseekr1}, and Gemini 2.5 \citep{GeminiTeam2025gemini25} use chain-of-thought reasoning combined with reinforcement learning and search \citep{Bertietal2025,Weietal2022,Zhangetal2025}. These models excel at reasoning but rely on static knowledge \citep{Lietal_12025,Lietal_22025,Huangetal2025}, which can lead to errors on knowledge-intensive tasks \citep{Ferragetal2025} and produce long reasoning traces that increase cost and latency \citep{Liuetal2025,Zhangetal2025}. Retrieval-augmented methods aim to address these limitations, though their effectiveness varies. RAG based methods  struggles with multi-hop reasoning, while Self-Ask \citep{Press2023SelfAsk}, ReAct \citep{Yao2023ReAct}, and IRCoT \citep{Trivedi2023IRCoT} interleave retrieval but often generate weak queries.. More recent approaches such as Search-o1 \citep{Lietal_12025}, EXSEARCH \citep{shi2025iterative}, Search-R1 \citep{jin2025search}, and WebThinker \citep{li2025webthinker} incorporate planning or uncertainty handling, but they still lack true multi-agent collaboration.

Multi-agent methods boost robustness through specialization and interaction. AgentVerse \citep{chen2023agentverse} coordinates agents, MACNET scales reasoning with DAGs \citep{qian2024scaling}, and consensus frameworks like Debate \citep{wang2025learning} and ReConcile \citep{chen2023reconcile} raise reliability. AutoTIR \citep{wei2025autotir} couples agents with reinforcement-based tool use, but most still lack \textbf{on-demand search and retrieval}. Instead of dynamically generating targeted queries (e.g., searching only when an agent encounters uncertainty in multi-hop reasoning), or flexibly grounding results in richer persona-specific or hypothetical documents (e.g., HyDE \citep{Gao2023HyDE}), they continue to rely on traditional lexical or embedding-based matches and fixed, upfront toolchains.

\noindent
\textbf{Mathematical reasoning in LLMs:} Recent work has also improved LLMs' ability to solve mathematical problems without relying on retrieval or multi-agent strategies. Chen et al.\ \citep{chen2024large} train models on the DEMI-MathAnalysis dataset to generate rigorous proofs in limits and series. Zhang et al.\ \citep{zhang2024llamaberry} introduce LLaMA-Berry, which combines Self-Refine, Monte Carlo Tree Search, and a preference model to explore and rank reasoning paths. These works show that model-based reasoning can improve problem-solving without external knowledge. Lin et al.\ \citep{lin2025step} propose Step-kto, employing binary stepwise feedback to enhance reasoning efficiency. However, these approaches do not incorporate \textbf{on-demand knowledge integration}, which is essential when equations require targeted retrieval under uncertainty, nor do they utilize \textbf{multi-agent strategies}, where specialized solver agents collaborate to improve robustness.

\begin{figure*}[t]
    \centering
    \vspace{1em} 
    \includegraphics[width=\textwidth]{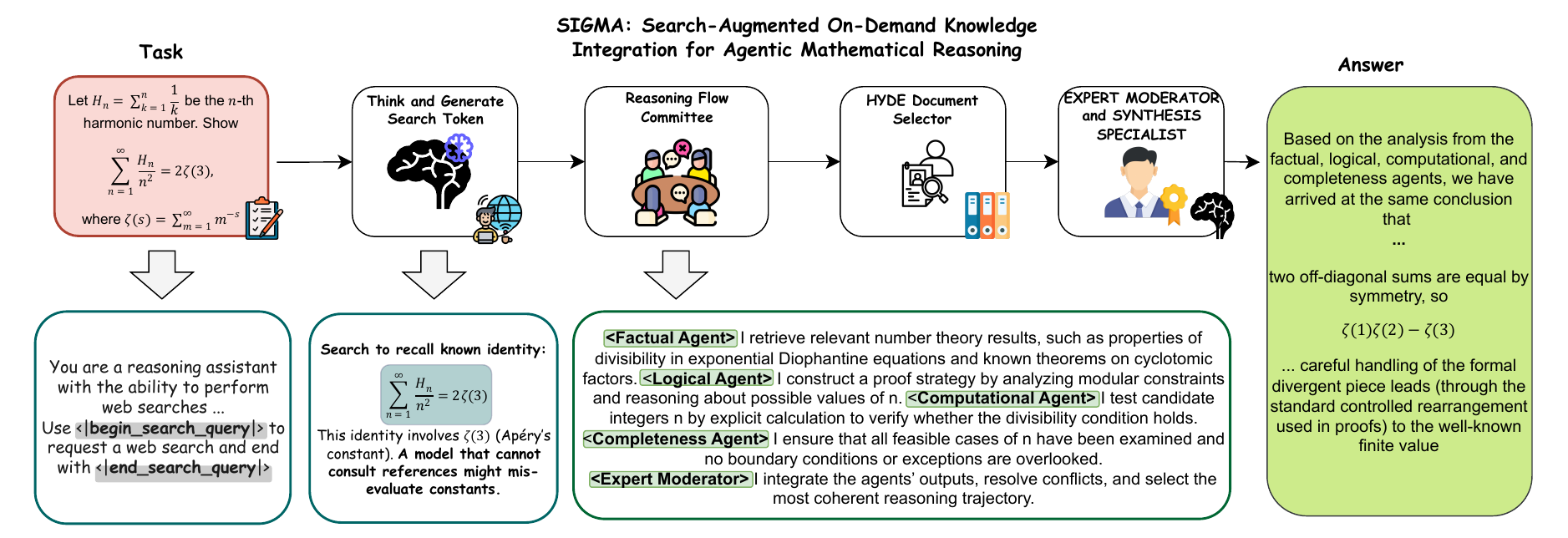}
    \vspace{-1.3 em} 
    \caption{Overview of the SIGMA framework.}
    \label{fig:method-figure}
    \vspace{-0.3em} 
\end{figure*}

\begin{table*}[ht!]
\centering
\caption{Main results on challenging reasoning tasks, including Maths and PhD-level science QA. We report Pass@1 metric for all tasks. The best results are in \textbf{bold} and the second-best are \underline{underlined}. Symbol ``$\dagger$'' indicates results from their official releases.}
\label{tab:main-results}
\resizebox{\textwidth}{!}{
\begin{tabular}{l|ccc|cccc}
\toprule
\multicolumn{1}{c|}{\textbf{Method}} & \multicolumn{3}{c|}{\textbf{Math Benchmarks}} & \multicolumn{4}{c}{\textbf{GPQA (PhD-Level Science QA)}} \\
 & MATH500 & AMC23 & AIME24 & Physics & Chemistry & Biology & Overall \\
\midrule
\multicolumn{8}{l}{\textit{Direct Reasoning (w/o Retrieval)}} \\
\textit{Small-Scale Models (\ensuremath{<8}B Parameters)} & & & & & & & \\
Qwen2.5-3B & 57.4 & 50.0 & 3.33 & 32.56 & 26.88 & 68.42 & 33.33 \\
Qwen2.5-7B & 64.2 & \underline{62.5} & 16.67 & 37.20 & 26.88 & 42.10 & 32.83 \\
Gemini 1.5 Flash-8B \citep{GeminiTeam2025gemini25}& 67.8 & - & - & - & - & - & - \\
\textit{Large-Scale Models (\ensuremath{>32}B Parameters)} & & & & & & & \\
\textcolor{gray}{Claude-3-5-Haiku} \citep{anthropic_claude_3_5_haiku_2024} & \textcolor{gray}{67.0} & \textcolor{gray}{-} & \textcolor{gray}{5.3} & \textcolor{gray}{-} & \textcolor{gray}{-} & \textcolor{gray}{-} & \textcolor{gray}{41.6} \\
\textcolor{gray}{Qwen2.5-Coder-32B} & \textcolor{gray}{71.2} & \textcolor{gray}{67.5} & \textcolor{gray}{20.0} & \textcolor{gray}{37.2} & \textcolor{gray}{25.8} & \textcolor{gray}{57.9} & \textcolor{gray}{33.8} \\
\textcolor{gray}{Llama3.3-70B} \citep{llama3} & \textcolor{gray}{70.8} & \textcolor{gray}{47.5} & \textcolor{gray}{36.7} & \textcolor{gray}{54.7} & \textcolor{gray}{31.2} & \textcolor{gray}{52.6} & \textcolor{gray}{43.4} \\
\textcolor{gray}{GPT-4o$\dagger$} \citep{gpt_4o_system_card} & \textcolor{gray}{60.3} & \textcolor{gray}{-} & \textcolor{gray}{9.3} & \textcolor{gray}{59.5} & \textcolor{gray}{40.2} & \textcolor{gray}{61.6} & \textcolor{gray}{50.6} \\
\midrule
\multicolumn{8}{l}{\textit{Custom Training Methods}} \\
STEP-KTO M3 (Llama-3.1-8B-Instruct) \citep{lin2025step} & 63.2 & 47.5 & 16.7 & - & - & - & - \\
\midrule
\multicolumn{8}{l}{\textit{Tool Integrated Methods}} \\
Search-o1 (Qwen2.5-1.5B) \citep{li2025search} & 36.0 & 25.0 & 3.33 & 27.9 & 17.2 & 0.0 & 20.2 \\
\textbf{SIGMA (Ours)} (Qwen2.5-1.5B) & 43.4 & 20.0 & 6.66 & 23.26 & 16.12 & 15.79 & 19.19 \\
Search-o1 (Qwen2.5-3B) \citep{li2025search} & 58.2 & 50.0 & 6.66 & 18.6 & 24.73 & 15.79 & 21.21 \\
\textbf{SIGMA (Ours)} (Qwen2.5-3B) & 61.0 & 30.0 & 6.66 & \underline{34.88} & \textbf{29.03} & \underline{63.16} & \underline{34.85}\\
Auto-TIR (Qwen2.5-7B) \citep{wei2025autotir} & 62.6 & - & \textbf{33.33} & - & - & - & - \\
Search-o1 (Qwen2.5-7B) \citep{li2025search} & 64.8 & 55.0 & 13.33 & 27.9 & 24.73 & 63.16 & 29.79 \\
\textbf{SIGMA (Ours)} (Qwen2.5-7B) & \textbf{68.4} & \textbf{60.0} & \underline{16.67} & \textbf{37.2} & \underline{27.96} & \textbf{68.42} & \textbf{35.86} \\
\bottomrule
\end{tabular}
}
\end{table*}\vspace{-0.6em} 

\section{Methodology}

We propose the \textbf{SIGMA framework} (see Figure \ref{fig:method-figure}) for complex reasoning tasks requiring external knowledge through dynamic multi-agent orchestration within a unified model $M_\theta$. Given a query $q$, for example \emph{find all positive integers $n$ such that $2^n + 1$ divides $3^n - 1$}, SIGMA integrates knowledge on demand, retrieving relevant theorems, strategies, computational tools, and edge cases as needed. To facilitate external knowledge access, the model is explicitly instructed to use the special token {\color{blue}\texttt{<|begin\_search\_query|>}} when a web search is required, enabling targeted retrieval of supporting information (see Appendices \textbf{A} and \textbf{B}).

Unlike conventional Tool-Integrated Reasoning \citep{li2025webthinker, jin2025search, shi2025iterative, Lietal_12025} with rigid \textbf{\textit{think-action-observation}} workflows, SIGMA enables flexible collaboration among four specialized agents: \textsc{Factual}, \textsc{Logical}, \textsc{Computational}, and \textsc{Completeness}. Each agent follows an independent reasoning trajectory and autonomously invokes searches. For example, \textsc{Factual} may retrieve results from number theory, \textsc{Logical} considers proof strategies, \textsc{Computational} verifies candidates, and \textsc{Completeness} checks boundary cases. Agents maintain persistent states, which evolve over time as follows:
\begin{equation}
S_k^t = f_\theta(S_k^{t-1}, \phi_k^{t-1}, o_k^{t-1}),
\quad \phi_k^t \sim P(\phi \mid S_k^t)
\end{equation}
where $S_k^t$ represents the internal state of agent $k$ at reasoning step $t$. This state is updated by the unified model function $f_\theta$ (parameterized by $\theta$) based on its previous state $S_k^{t-1}$, the action $\phi_k^{t-1}$ taken at the previous step, and the observation $o_k^{t-1}$ (e.g., a search result or internal thought) resulting from that action. The agent's next action $\phi_k^t$ is sampled from a policy $P(\phi \mid S_k^t)$, which is conditioned on the current state. Actions are drawn from a discrete set $\phi_k^t \in \{\phi_{\text{search}}, \phi_{\text{reason}}, \phi_{\text{synthesize}}\}$, allowing an agent to autonomously decide whether to search, reason internally, or synthesize its findings. A final moderator then synthesizes outputs across all agents $\mathcal{A}$ to produce the final answer $a_{\text{final}}$:
\begin{equation}
a_{\text{final}} = \text{Moderator}(\{S_k^T, a_k\}_{k \in \mathcal{A}})
\end{equation}
Here, $\{S_k^T, a_k\}$ represents the set of final states and intermediate answers generated by each agent $k$ upon reaching its terminal step $T$.

To clarify the moderator's role, the \textbf{moderator} component is implemented as a non-learnable, heuristic synthesis layer rather than a trained module. Its primary function is to integrate the distinct outputs $\{a_k\}$ from each agent. To handle redundant outputs, the moderator first collates and deduplicates all unique propositions. In cases of conflicting reasoning, it employs a pre-defined prioritization scheme; for example, verified results from the \textsc{Computational} agent are given higher weight than speculative hypotheses from the \textsc{Logical} agent. This deterministic approach ensures consistent synthesis while maintaining focus on the agents' reasoning capabilities. By modeling collaboration within a unified inference process, SIGMA achieves contextual continuity and computational efficiency. 

\noindent
\textbf{Hypothetical document enhancement:} To support high-quality, targeted retrieval, each agent $k$ generates hypothetical passages using previous work HyDE \cite{Gao2023HyDE} $P_{\text{hyde}}^{(k,j)}$ representing ideal answer content given its current reasoning state.
\begin{equation}
P_{\text{hyde}}^{(k,j)} = \text{HypoGenerator}(q, S_k^j, q_j^{(k)})
\end{equation}
This generation is guided by the original query $q$, the agent's specific state $S_k^j$ at its $j$-th search step, and any agent-specific sub-query $q_j^{(k)}$ it has formed. Candidate chunks $c_m^{(k,j)}$ retrieved from the external corpus are then ranked by their embedding similarity to this hypothetical document:
\begin{equation}
\text{Sim}_m^{(k,j)} = \cos(\text{Embed}(P_{\text{hyde}}^{(k,j)}), \text{Embed}(c_m^{(k,j)}))
\end{equation}
where $\text{Sim}_m^{(k,j)}$ is the cosine similarity score for the $m$-th candidate chunk. This ensures that retrieved information is precisely tailored to the agent's immediate needs while preserving coherence across the unified reasoning process.

A key design feature of SIGMA is the close integration of its specialized agents. Their reasoning processes are interconnected within the shared model $M_\theta$, allowing them to influence each other’s behavior (for example, \textsc{Completeness} questioning an assumption made by \textsc{Logical}, prompting a new search by \textsc{Factual}). Because of this tight coupling, removing a single agent (e.g., “SIGMA w/o \textsc{Factual}”) does not simply isolate its effect, it changes how the entire system functions. So, we excluded such ablations, as they would yield misleading results. Instead, our experiments evaluate the full SIGMA system against direct reasoning and tool-integrated baselines. Table \ref{tab:main-results} includes a qualitative analysis of how the agents interact during reasoning.

\section{Experiments}
\noindent
\textbf{Experimental Setup and Benchmarks.} We evaluate SIGMA’s reasoning capabilities on three mathematical benchmarks, MATH500 \citep{hendrycks2021measuring}, AMC2023, and AIME2024 \citep{tongyx361AwesomeLLM4Math}. MATH500 comprises 500 problems from the MATH test set \citep{hendrycks2021measuring}, while AMC2023 and AIME2024 represent middle school–level competitions covering arithmetic, algebra, and geometry. To assess domain-specific scientific reasoning, we additionally employ GPQA \citep{2311_gpqa}, a PhD-level multiple-choice dataset across physics, chemistry, and biology. Our experiments use the diamond subset to ensure high-quality evaluation. All baseline models were evaluated under identical search budgets, retrieval protocols, and decoding parameters (temperature, seed, pass@k) to ensure fair cross-model comparison.

\noindent
\textbf{Baseline and Model Selection.} We assess SIGMA as a \textit{general reasoning framework} applied to base foundation models, using \texttt{Qwen2.5-1.5B/3B/7B} \citep{qwen2.5} as backbones. This setup enables direct comparison with general-purpose tool-integrated approaches such as Search-o1 \citep{li2025search}. While specialized models like \texttt{Qwen2.5-Math-1.5B} \citep{qwen2.5math} demonstrate strong performance on MATH500 through task-specific fine-tuning, our focus is on evaluating SIGMA’s architectural contributions, specifically, its multi-agent orchestration and adaptive knowledge integration under a zero-shot setting. To maintain consistency with comparable baselines, we therefore employ the Qwen2.5 series, with integration of newer Qwen 3 models reserved for future exploration.

\textbf{Inference Cost and Efficiency.} SIGMA introduces a modest computational overhead relative to single-agent approaches such as direct reasoning or ReAct \cite{yao2022react}, as it runs up to four coordinated reasoning trajectories followed by a synthesis step. These trajectories interact within the shared model $M_\theta$, where specialized agents influence and refine each other's reasoning. Despite this coupling, the framework remains efficient because communication occurs implicitly through the shared reasoning space, avoiding the heavy coordination costs found in typical multi-agent systems. Although a detailed analysis of latency and token usage remains future work, SIGMA demonstrates consistent improvements across mathematical and reasoning benchmarks, as described in the Results section, while maintaining parallelizable and robust reasoning performance.

\section{Results}

Across all benchmarks, SIGMA consistently improves over both direct reasoning and prior tool-integrated baselines. On mathematical tasks, SIGMA (7B) achieves an absolute gain of \textbf{3.6\% }over Search- \citep{li2025search} at the same scale and\textbf{ 5.8}\% over Auto-TIR \citep{wei2025autotir} on MATH500. Compared to larger closed-source systems, SIGMA surpasses GPT-4o \citep{gpt_4o_system_card} by\textbf{ 8.1\%} on MATH500 and improves upon Claude-3.5-Haiku \citep{anthropic_claude_3_5_haiku_2024} by \textbf{1.4\%}, while coming close to Llama-3.3-70B \citep{llama3} despite being more than \textbf{10x smaller}. On AMC23 and AIME24, SIGMA further improves over retrieval-augmented baselines by \textbf{5.0\%} and \textbf{3.3\% }respectively, with particularly strong benefits on multi-step reasoning problems. Beyond mathematics, SIGMA generalizes effectively to GPQA, where it delivers a \textbf{6.1\%} overall gain relative to retrieval baselines, including a \textbf{9.3\%} improvement in Physics, \textbf{3.2\% }in Chemistry, and \textbf{5.3\%} improvement in Biology. These findings show that multi-perspective reasoning with moderator-based synthesis enables SIGMA to establish a new performance frontier across diverse reasoning domains.

\section{Conclusion}
We introduced SIGMA, a retrieval-augmented reasoning framework that coordinates mathematical problem-solving through specialized multi-agent perspectives. This approach enables comprehensive analysis through reasoning-search cycles and moderator-based synthesis. Experiments on MATH500, AIME, and GPQA show consistent improvements over existing baselines, demonstrating that multi-perspective reasoning with on-demand knowledge integration enhances accuracy in complex mathematical domains.

\bibliography{aaai2026}

\appendix

\section{SIGMA Multi-Agent Inference}\label{app:sigma}

\begin{algorithm}
\fontsize{9.6pt}{10.8pt}\selectfont
\caption{SIGMA Multi-Agent Inference}\label{algo:sigma_inference}
\begin{algorithmic}[1]
\Require Model $M_\theta$, search function \texttt{Search}, HDS selector
\State \textbf{Input:} Query $q$, agent instructions $\{I_k\}_{k \in \mathcal{A}}$
\For{each agent $k \in \{\textsc{Factual}, \textsc{Logical}, \textsc{Computational}, \textsc{Completeness}\}$}
    \State Initialize state: $S_k^0 \gets I_k \oplus q$
    \State Initialize search budget: $B_k \gets \text{MaxSearches}$
    \While{$B_k > 0$ and reasoning incomplete}
        \State Generate step: $r_k \gets M_\theta(S_k^t)$
        \State Update: $S_k^{t+1} \gets S_k^t \oplus r_k$
        \If{$r_k$ contains query $q_j^{(k)}$}
            \State \textbf{Emit:} \texttt{<|begin\_search\_query|>} $q_j^{(k)}$ \texttt{<|end\_search\_query|>}
            \State $\mathcal{D}_j^{(k)} \gets \texttt{Search}(q_j^{(k)})$
            \State $P_{\text{hyde}} \gets \text{HypoGenerator}(q, S_k^t, q_j^{(k)})$
            \State Apply HDS and integrate results
            \State $B_k \gets B_k - 1$
        \EndIf
    \EndWhile
    \State Extract conclusion: $a_k \gets \text{ExtractAnswer}(S_k^T)$
\EndFor
\State \textbf{Moderator Synthesis:} $a_{\text{final}} \gets \text{Moderator}(q, \{(S_k^T, a_k)\}_{k \in \mathcal{A}})$
\State \textbf{Output:} Final answer $a_{\text{final}}$
\end{algorithmic}
\end{algorithm}

\subsection{Problem and Multi-Agent Execution}
Given the query "Find positive integers $n \leq 2024$ such that $\gcd(n, 2024) = 1$", SIGMA processes this through four specialized agents with search budgets of 2 each. The FACTUAL agent begins by recognizing this as a number theory problem requiring verification of mathematical definitions, so it searches {\color{blue}\texttt{<|begin\_search\_query|>}} Euler totient function definition {\color{blue}\texttt{<|end\_search\_query|>}} and discovers that the problem asks for $\phi(2024)$ where $\phi(n)$ counts integers coprime to $n$. The agent then searches for prime factorization methods and concludes that computing $\phi(2024)$ requires factoring 2024 first. Meanwhile, the COMPUTATIONAL agent focuses on the mathematical calculations needed, searching {\color{blue}\texttt{<|begin\_search\_query|>}} factor 2024 prime decomposition {\color{blue}\texttt{<|end\_search\_query|>}} to find $2024 = 2^3 \times 11 \times 23$, then searching for totient calculation methods to compute $\phi(2024) = 2024 \times (1-\frac{1}{2}) \times (1-\frac{1}{11}) \times (1-\frac{1}{23}) = 880$. The LOGICAL agent establishes the conceptual framework by searching for the relationship between coprimality and Euler's function, confirming that $\gcd(n, 2024) = 1$ means $n$ shares no prime factors with 2024, which is exactly what the totient function counts. The COMPLETENESS agent verifies the solution by searching for alternative calculation methods and cross-validation approaches to ensure no steps were missed. After each agent completes its reasoning and search process, the moderator synthesizes their contributions: the FACTUAL agent provided the correct mathematical framework, the COMPUTATIONAL agent performed accurate calculations, the LOGICAL agent confirmed the conceptual connection, and the COMPLETENESS agent validated the approach. The final synthesis integrates these insights to produce: "The question asks for positive integers coprime to 2024, which equals Euler's totient function $\phi(2024)$. Using prime factorization $2024 = 2^3 \times 11 \times 23$ and the totient formula, we get $\phi(2024) = 880$."

\subsection{Framework Advantages}
The multi-agent approach succeeds where single-agent systems often struggle because it distributes different types of mathematical reasoning across specialized agents. The FACTUAL agent ensures definitional accuracy, preventing conceptual errors. The COMPUTATIONAL agent handles numerical calculations systematically. The LOGICAL agent maintains structural coherence. The COMPLETENESS agent catches potential oversights. This distributed expertise, combined with targeted search queries from each agent's perspective, creates more robust solutions for complex mathematical problems requiring both theoretical understanding and computational precision.
%%%%%%%%%%%%%%%%%%%%%%%%%%%%%%%%%%%%%%%%%%%%%%%%%%%%%%%%%%%%

\end{document}